%% file: paper.tex
\documentclass[twoside, 11pt]{article}
\usepackage{jmlr2e}

\input{def.tex}

\begin{document}
	
\title{A Variant of the Wang-Foster-Kakade Lower Bound for the Discounted Setting}

\author{\name Philip Amortila \email  philipa4@illinois.edu\\
\name Nan Jiang \email nanjiang@illinois.edu \\
\name Tengyang Xie \email tx10@illinois.edu\\
\addr University of Illinois at Urbana-Champaign
}

\maketitle

\newcommand{\para}[1]{\paragraph{#1}}

\renewcommand*{\theHsection}{\thesection}
\renewcommand*{\theHsubsection}{\thesubsection}

\begin{abstract}
	Recently, \citet{wang2020statistical} showed a highly intriguing hardness result for batch reinforcement learning (RL) with linearly realizable value function and good feature coverage in the finite-horizon case. In this note we show that once adapted to the discounted setting, the construction can be simplified to a $2$-state MDP with $1$-dimensional features, such that learning is impossible even with an infinite amount of data. 
\end{abstract}

\vspace*{1em} \noindent
\citet{wang2020statistical} recently showed that in finite-horizon batch RL, the sample complexity of evaluating a given policy $\pi$ has an information-theoretic lower bound that is exponential in the horizon, even if  realizable linear features  are given (i.e., $\varphi:\Scal\times\Acal\to\RR^d$ such that $Q^\pi(\cdot) = \langle \varphi(\cdot), \theta^\pi \rangle$ for some parameter $\theta^\pi \in \mathbb{R}^d$) and data provides good feature coverage (i.e., $\EE[\varphi \varphi^\top]$ has lower-bounded eigenvalues under the data distribution).  In this document we show that its analogy in the discounted setting has  a stronger statement  (\emph{infinite} sample complexity) with a simpler construction (1-d feature, $2$ states, and arbitrary discount factor.)

\begin{figure}[h]
	\centering
	\includegraphics[scale=1.0]{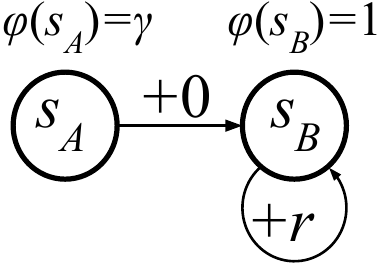}
	\caption{Construction for $d=1$. \label{fig:lower}}
\end{figure}

Consider the deterministic MDP in Figure~\ref{fig:lower} with a discount factor $\gamma \in (0, 1)$, where every state only has 1 action (which we omit in the notations). $s_A$ transitions to $s_B$ with $0$ reward, and $s_B$ has a self-loop with $r$ reward per step. The batch data only contains the tuple $(s_A, 0, s_B)$. The feature map is 1-dimensional: $\varphi(s_A) = \gamma$ and $\varphi(s_B) = 1$. Clearly, without data from $s_B$, the learner cannot know the value of $r$, hence cannot determine the value of $s_A$ or $s_B$, even with an infinite amount of data.

We now verify the realizability and coverage assumptions: 
\begin{itemize}
\item \textbf{Realizability:} We show that $V^\pi(\cdot) = \langle \varphi(\cdot), \theta \rangle$ for some $\theta \in \mathbb{R}$. By the Bellman equation, $V^\pi(s_A) = \gamma V^\pi(s_B)$. Therefore, $V^\pi(s_A) = \langle \varphi(s_A), V^\pi(s_B) \rangle$. Similarly, $V^\pi(s_B) = 1 \cdot V^\pi(s_B) = \langle \varphi(s_B) , V^\pi(s_B) \rangle$. So $V^\pi(\cdot)$ is always linearly-realizable, with $\theta = V^\pi(s_B) = \frac{r}{1-\gamma}$ being the unknown coefficient. 
\item \textbf{Coverage:} Translating the condition of  \citet{wang2020statistical} to the discounted case, it is required that: (1) $\|\varphi(\cdot)\|_2 \le 1$ always holds, and (2) $\EE[\varphi\varphi^\top]$ has polynomially lower bounded eigenvalues. (1) is satisfied in our construction. For (2), since we only have data from $s_A$, the feature covariance matrix under the data distribution is $\varphi(s_A) \varphi(s_A)^\top = \gamma^2$, whose only eigenvalue is $\gamma^2$ and is well above $0$ as long as $\gamma$ is. 
\end{itemize}

\paragraph{Extensions for general $d$ and the controlled setting} We briefly sketch two extensions of the construction. Although it is sufficient to prove the lower bound for $d=1$, the construction easily scales to arbitrary $d$: we simply make $d$ copies of the construction in Figure~\ref{fig:lower}, and assign a coordinate of $\varphi: \Scal\to\RR^d$ to each copy. Let data be uniform over the $s_A$ of all copies, so the feature covariance matrix is $\gamma^2/d \cdot I$. 

The extension to the controlled case is similar. Let $a$ denote the action of $s_A$ in Figure~\ref{fig:lower}. We introduce a second action $a'$ for $s_A$ that transitions to $s_C$ with $0$ reward, and $s_C$ is absorbing with reward $r'$. Let the 2-dimensional feature map be: $\varphi(s_A,a) = [\gamma, 0]^\top$, $\varphi(s_A,a') = [0, \gamma]^\top$, $\varphi(s_B) = [1, 0]^\top$, $\varphi(s_C) = [0, 1]^\top$. It is easy to verify that $Q^\star$ is realizable\footnote{In fact, Q-functions in this MDP do not depend on the policy, since only $s_A$ has multiple actions.}, but $Q^\star(s_A, a) = \frac{\gamma}{1-\gamma} r$ and $Q^\star(s_A, a') = \frac{\gamma}{1-\gamma} r'$ can independently take arbitrary values between $[0, \gamma/(1-\gamma)]$ (assuming rewards lie in $[0, 1]$), so the learner cannot choose a near-optimal action even with infinite data. \\

These observations combine to give us the following result:

\begin{proposition}[Informal]
For any $d \geq 1, \gamma \in (0,1)$, 
given realizable linear features, the value function learned by any batch RL algorithm must have $\Omega(1)$ worst-case error, even with an infinitely large dataset that has $\Theta(1/d)$ feature coverage. 
\end{proposition}

\paragraph{Final Remark} While the discounted setting allows a very simple construction for the lower bound, this does not imply that the construction for the finite-horizon setting can be simplified in a similar manner. In fact, we believe that the careful construction of \citet{wang2020statistical} that cleverly exponentiates a negligibly small error is necessary for the finite-horizon setting. Such a difference between the finite-horizon setting and the discounted setting, however, does challenge the conventional wisdom that the results in the finite-horizon setting and the discounted setting are often similar and translate to each other with $H=O(1/(1-\gamma))$ up to minor differences. Are these two lower bounds ``essentially the same'', or does their difference imply some fundamental difference between the finite-horizon and the discounted settings? We leave this open question to the readers.

\section*{Acknowledgement}
NJ thanks Ruosong Wang for helpful discussions. 

\bibliography{bib}
\end{document}

%% file: def.tex
\usepackage{url, hyperref}
\usepackage{algorithm, algpseudocode} %

\usepackage{amsthm, amsmath, amssymb} 
\usepackage[shortlabels]{enumitem}
\usepackage{comment}
\usepackage{color}
\usepackage{multirow}
\usepackage{xspace}
\usepackage{rotating}
\usepackage{graphicx}
\usepackage{caption}
\usepackage{xcolor}
\usepackage[normalem]{ulem}
\usepackage{tikz}

\usepackage{nicefrac}
\usepackage{mathtools}
\mathtoolsset{showonlyrefs,showmanualtags}

\theoremstyle{plain}
\newtheorem{theorem}{\textbf{Theorem}}

\newtheorem{proposition}[theorem]{Proposition}

\theoremstyle{definition}

\renewcommand{\cite}{\citep}

\newcommand{\EE}{\mathbb{E}}

\newcommand{\Scal}{\mathcal{S}}

\newcommand{\Acal}{\mathcal{A}}

\newcommand{\RR}{\mathbb{R}}

